\documentclass[conference]{IEEEtran}
\usepackage[bookmarks=false,hidelinks=true]{hyperref}
\usepackage{graphicx}
\usepackage{amsmath}
\usepackage{amssymb}
 
\usepackage{float}
\usepackage{listings}
\usepackage{color}
\usepackage{xcolor}
\usepackage{caption}
\usepackage{subfigure}
\usepackage{amsmath}
\usepackage[linesnumbered,ruled]{algorithm2e}
\usepackage[noend]{algpseudocode}
\usepackage{epigraph}
\begin{document}

\title{\LARGE \bf
A fully end-to-end deep learning approach for real-time \\
simultaneous 3D reconstruction and material recognition}

\author{Cheng Zhao, Li Sun and Rustam Stolkin \\
Extreme Robotics Lab, University of Birmingham, UK.\\
IRobotCheng@gmail.com
}

\maketitle

\begin{abstract}
This paper addresses the problem of \textit{simultaneous} 3D reconstruction and material recognition and segmentation. Enabling robots to recognise different materials (concrete, metal etc.) in a scene is important for many tasks, e.g. robotic interventions in nuclear decommissioning. Previous work on 3D semantic reconstruction has predominantly focused on recognition of everyday domestic objects (tables, chairs etc.), whereas previous work on material recognition has largely been confined to single 2D images without any 3D reconstruction. Meanwhile, most 3D semantic reconstruction methods rely on computationally expensive post-processing, using Fully-Connected Conditional Random Fields (CRFs), to achieve consistent segmentations. In contrast, we propose a deep learning method which performs 3D reconstruction while simultaneously recognising different types of materials and labeling them at the pixel level. Unlike previous methods, we propose a fully end-to-end approach, which does not require hand-crafted features or CRF post-processing. Instead, we use only learned features, and the CRF segmentation constraints are incorporated inside the fully end-to-end learned system. We present the results of experiments, in which we trained our system to perform real-time 3D semantic reconstruction for 23 different materials in a real-world application. The run-time performance of the system can be boosted to around 10Hz, using a conventional GPU, which is enough to achieve real-time semantic reconstruction using a 30fps RGB-D camera. To the best of our knowledge, this work is the first real-time end-to-end system for simultaneous 3D reconstruction and material recognition.    
\end{abstract}

\begin{IEEEkeywords}
3D semantic reconstruction, material recognition, real-time, fully end-to-end, deep neural network 
\end{IEEEkeywords}
\section{Introduction}
Real-time 3D semantic reconstruction is required in many robotics applications, such as autonomous navigation or grasping and manipulation. While a variety of well-known methods \cite{Endres2014}\cite{Engel2014}\cite{Newcombe2011} can reconstruct accurate 3D maps at real-time frame rates, the resulting point-clouds contain no semantic-level understanding of the observed scenes. Hence, the problem of 3D semantic reconstruction is attracting increasing attention in the robotics research community. Recent methods \cite{Salas-Moreno2013}\cite{Tateno2016}\cite{Hermans2014}\cite{Prisacariu2015}\cite{McCormac2016} not only generate a 3D point cloud map, but also simultaneously assign a semantic label to each point in the cloud. However, these methods are typically designed to search for everyday domestic 3D objects  (e.g. ``table'', ``chair'' etc.) in domestic (non-industrial) scenes.

\begin{figure}[thpb]
	\centering
	\includegraphics[width=0.5\textwidth]{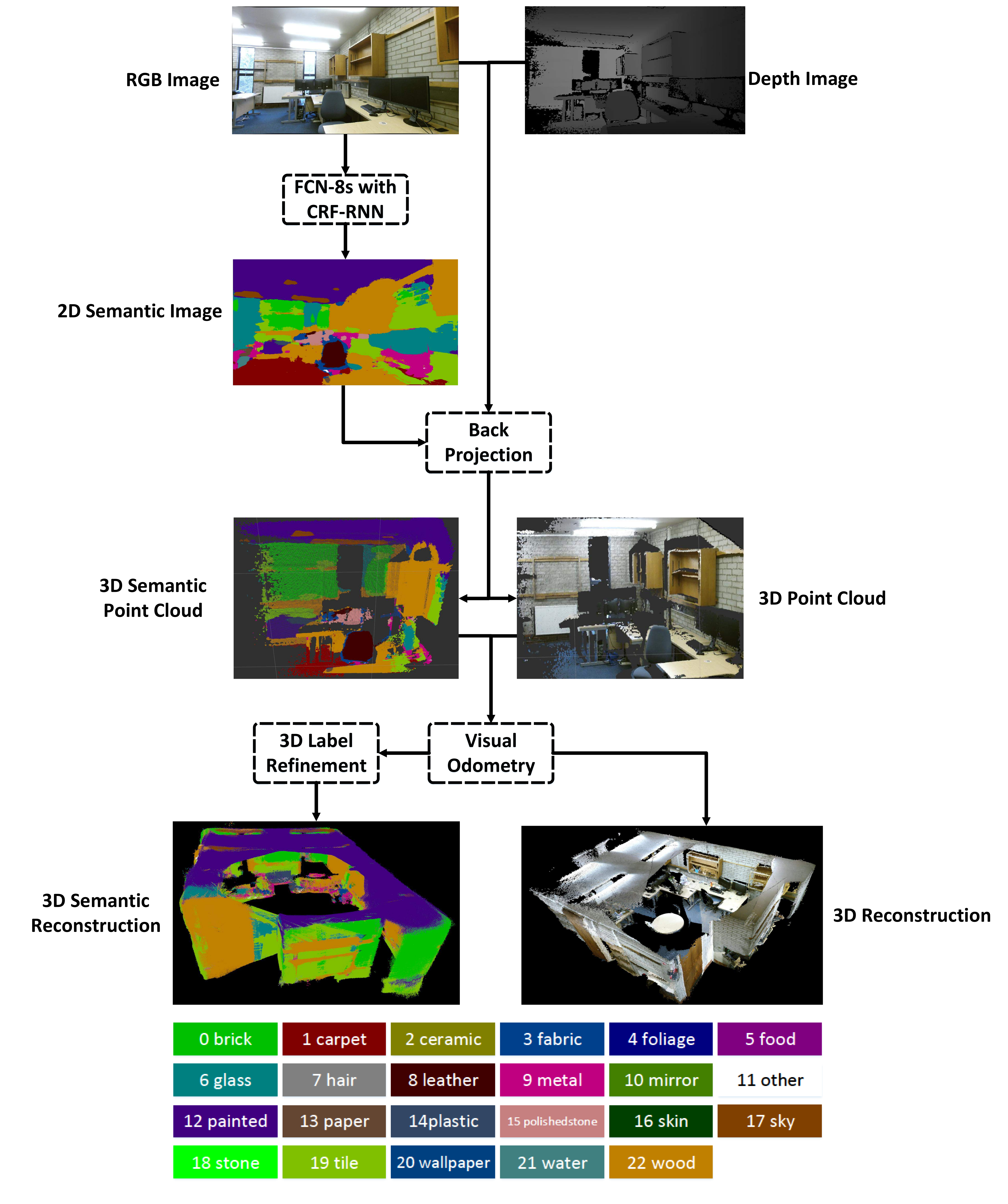}
	\caption{\textbf{Pipeline of proposed simultaneous 3D reconstruction and material recognition system}. Firstly, FCN-8s with CRF-RNN is employed for 2D material recognition using the RGB image from RGB-D camera. Then the semantically labeled RGB image, and the corresponding depth image, are combined together through back-projection to generate a semantic point cloud for each key frame. Finally, all semantic point clouds are combined incrementally using visual odometry, and Bayesian update is employed for label probability refinement.}
	\label{fig:pipeline_3D_semantic_reconstruction}
\end{figure}

In contrast, an ability to recognize different kinds of \textit{materials} could play a very important role in numerous robotics applications. Understanding material properties (e.g. friction or deformability) of objects can be used to inform grasp planning and manipulation. Rescue robots should understand their surrounding materials when planning movements through precarious rubble under a collapsed building. In nuclear decommissioning, robots must enter hazardous zones (in very old legacy buildings with significant uncertainty) to perform ``characterisation''. Understanding which materials make up the scene is a critical aspect of characterisation, and will inform subsequent interventions, e.g. cutting, dismantling, cleaning, manipulating. Unfortunately, previous work \cite{Degol2016}\cite{Bell2015}\cite{Schwartz2016}\cite{Rusk2016} only addresses the problems of detecting and segmenting materials in a single RGB image, and does not perform 3D material reconstruction.


In this paper, we present a fully end-to-end system, which performs real-time 3D reconstruction while simultaneously recognizing and labeling each pixel according to its material, Fig.\ref{fig:pipeline_3D_semantic_reconstruction}. The main contributions of this paper can be summarized as follows:
\begin{itemize}
\item To the best of our knowledge, this is the first system to perform simultaneous 3D reconstruction and material recognition.
\item Hand-crafted features or post-processing CRF optimization are not required. In contrast, the system is fully end-to-end learned, and this helps to deliver real-time performance, as well as generality for different applications.
\item The run-time performance of the whole system can be boosted, using a conventional GPU, to around 10Hz, which is enough to achieve real-time semantic reconstruction using a 30fps RGB-D camera.
\item We demonstrate our method in a real-world application, reconstructing a room while simultaneously recognizing and labeling 23 different materials. 
\end{itemize}  

\section{Related Work}
In this section, we firstly review real-time 3D semantic reconstruction in Section \ref{sec:Real-time 3D semantic reconstruction} and material recognition in Section \ref{sec:Material recognition}. Then we will give a discussion in Section \ref{sec:Discussion}. 

\subsection{Real-time 3D semantic reconstruction}\label{sec:Real-time 3D semantic reconstruction}
Recent real-time 3D reconstruction and SLAM approaches, e.g. \cite{Endres2014} (Visual-features with RANSAC), \cite{Engel2014} (Direct image alignment based on optimization),  or \cite{Newcombe2011}(Point cloud alignment based on ICP) can effectively generate dense or semi-dense 3D maps, but they have no understanding of the observed scenes and objects. 

The more complex problem of 3D semantic reconstruction remains an open research problem. Recent approaches can be grouped into two main categories: 3D semantic reconstruction base on 3D template matching \cite{Salas-Moreno2013}\cite{Salas-Moreno}\cite{Tateno2016}, and 3D semantic reconstruction base on 2D semantic segmentation \cite{Hermans2014}\cite{Prisacariu2015}.

The former methods rely on 3D template matching, so can only be used in situations with many repeated and identical objects. Only known 3D objects can be recognised, and semantic labeling for the remainder of the scene is not possible. The latter methods typically employ visual features, combined with a classifier such as random forest, for 2D semantic segmentation. The visual features are hand-crafted and also requires a non-linear transformation between local and global descriptors. Unlike CNN features, which can be learned from training data in an end-to-end fashion, such methods require application-specific, human-designed components. 

The work most closely related to ours is \cite{McCormac2016}, which performs dense, 3D semantic mapping ofindoor scenes, using deconvolutional neural networks\cite{Noh2016}. Real-time frame-rates of about 25 Hz are achieved \textit{without} CRF optimization post-processing. But most such methods \cite{Hermans2014}\cite{Prisacariu2015}\cite{McCormac2016} rely on using fully connected CRF\cite{Krahenbuhl2012} optimization as an offline post-processing step, following online 3D reconstruction, i.e. these methods do not actually achieve semantic mapping in real-time. Additionally, these methods are not fully end-to-end trainable, and there is no interaction between classifier learning and CRF learning. The parameters of classifier and CRF cannot be jointly learned in a united framework. Furthermore, all of the above methods are focused on semantic \textit{object} recognition. In contrast, our work is the first method that achieves simultaneous 3D reconstruction with semantic \textit{material} labeling, and we achieve both 3D reconstruction \textit{and} semantic labeling simultaneously in real-time.

\subsection{Material recognition}\label{sec:Material recognition}
Materials recognition is a challenging research topic due to wide variation in appearance within categories. Previous material recognition research predominantly focused on material classification, and did not achieve pixel-wise material segmentation. Most previous work employed hand-crafted visual features, e.g. reflectance-based edge features \cite{Liu2010}, variances of oriented gradients \cite{Hu2011}, and pairwise local binary patterns \cite{Qi2014}. Recently CNN features \cite{Schwartz2013}\cite{Vedaldi}\cite{Mcdonald-maier2016} have been employed to achieve the state-of-the-art results of material classification in many public material datasets. In addition to the 2D features, \cite{Degol2016} combined 3D geometry (surface normals, camera intrinsic and extrinsic parameters) with 2D features (texture and color) to improve material classification.

For pixel-wise material segmentation, \cite{Bell2015} convert patch-based trained CNN classifiers into an efficient fully convolutional framework combined with a fully connected CRF to perform pixel-wise material recognition. \cite{Schwartz2016} combined local appearance with separately recognized global contextual cues including objects and places, which can lead to a superior result. They employed fully convolutional network(FCN) \cite{Long2015} followed by recurrent neural network(RNN) for dense pixel-wise material segmentation. \cite{Rusk2016} proposed a novel CNN architecture trained on 4D light-field images and employ FCN for per-pixel material recognition. However, in contrast to our work, none of these methods perform material recognition simultaneously with 3D reconstruction. 

\subsection{Discussion}\label{sec:Discussion}
In summary, previous real-time semantic 3D reconstruction methods have focused on object recognition, and not on material recognition. Predominantly, such methods require post-processing with a fully-connected CRF, and are not fully end-to-end. While there is literature on per-pixel material recognition in 2D images, no previously reported methods perform 3D material reconstruction. To our best knowledge, this work is the first real-time end-to-end system for simultaneous 3D reconstruction and material recognition.    
\section{Methods}
\subsection{Overview}
The pipeline of simultaneous 3D reconstruction and material recognition comprises three units as illustrated in Figure \ref{fig:pipeline_3D_semantic_reconstruction}: a real-time 3D reconstruction unit based on RGB-D SLAM\cite{Endres2014}, a 2D material recognition unit based on FCN-8s\cite{Long2015} with CRF-RNN\cite{Zheng2015}, and a 3D semantic reconstruction unit based on Bayesian update. Firstly, the FCN-8s with CRF-RNN is employed for 2D material recognition using the RGB image from RGB-D camera. Then the semantically labeled RGB image, and the corresponding depth image, are combined together through back-projection to generate a semantic point cloud for each key frame. Finally, all semantic point clouds are combined incrementally using visual odometry, and Bayesian update is employed for label probability refinement.
\subsection{RGB-D SLAM Mapping}
We use the RGB-D SLAM method of \cite{Endres2014} for real-time 3D reconstruction. It is a graph-based SLAM system which includes a front-end system to processes the RGB-D sensor data to calculate geometric relationships through visual features based on RANSAC and ICP. Subsequently, the back-end system registers pairs of image frames to construct a pose graph. G2O\cite{Kummerle2011} is employed for graph optimization to obtain a maximum likelihood solution for the camera trajectory. Finally, RGB-D sensor data is combined together to generate a 3D point cloud.

In our system, RGB-D SLAM plays two important roles: 1) it can provide the transformation information between two adjacent semantic point clouds, enabling incremental semantic label fusion; 2) the visual odometry is used to combine all semantic point clouds to generate a global semantic map.  
\subsection{2D material recognition}
Our neural network is implemented in caffe \cite{Jia2014} framework and employs FCN-8s \cite{Long2015} followed by CRF-RNN \cite{Zheng2015} architecture. 

\subsubsection{FCN}
FCN is the first end-to-end and pixel-to-pixel semantic segmentation architecture which can take an input of arbitrary size and generate correspondingly-sized output images. This architecture is based on the VGG 16-layer\cite{Simonyan2015} network. The learned representations in the VGG-16 network can be transferred through fine-tuning our network, using the extracted patches in public material dataset MINC\cite{Bell2015}. Next, we transplant the fully connected VGG network into a fully convolutional VGG network and inherit the weights of the fine-tuning network. Finally, the FCN-32s, FCN-16s and FCN-8s networks are trained using MINC dataset sequentially. FCN defines a skip architecture which can combine semantic information from a deep, coarse layer with shape information from a shallow, fine layer. The output of the deep layer has rich semantic information but loses most of the shape information, while the shallow layer has rich shape information but lacks semantic information. Therefore FCN improve the accuracy of semantic segmentation by fusing the outputs from both deep and shallow layers. 

FCN has convolutional filters with large receptive fields and 5 pooling layers. It does not incorporate smoothness constraints between neighbouring pixels. Hence, it can only generate coarse pixel-wise predictions with blob-like shapes. In our system, FCN-8s is employed as the first part of the network to provide unary potentials to the CRF-RNN.
      
\subsubsection{CRF-RNN}
CRF-RNN, following FCN, combines the strengths of both FCN and fully-connected CRF into a single end-to-end unified framework. Fully-connected CRF accounts for contextual information by minimising the energy $E(x)$ function in the Gibbs distribution, to generate the most likely label assignment $x$. 

Energy function $E(x)$ consists of a unary data term and pairwise smoothness term, Equation \ref{eq:E(x)}. Unary term $\psi_{u}(x_{i})$ is obtained from the FCN-8s, which predicts pixel labels without considering smoothness or consistency of label assignments. The pairwise term $\psi_{p}(x_{i}, x_{j})$ encourages similar labeling of pixels with similar properties, while penalizing similar pixels which have different labels. 

\begin{equation}
E(x) = \sum_i \psi_{u}(x_{i}) + \sum_{i<j} \psi_{p}(x_{i}, x_{j})
\label{eq:E(x)}
\end{equation}

Pairwise potentials are modeled as a linear combination of $M$ Gaussian edge potential kernels as shown in equation \ref{eq:pairwise energy}. $f_i$ is the feature vector of pixel $i$ (e.g. spatial or colour information). $k_G^{(m)}$ is a Gaussian kernel applied to feature vectors. $\omega^{(m)}$ is the linear combination weight. The Potts model $\mu (x_i, x_j) = [ x_i \neq x_j] $ is the label compatibility function. 

\begin{equation}
\psi_{p}(x_{i}, x_{j}) = \mu (x_i, x_j)\sum^{M}_{m=1}\omega^{(m)}k^{(m)}_G(f_i, f_j)
\label{eq:pairwise energy}
\end{equation}

A bilateral appearance potential and a spatial smoothing potential($ M = 2 $) are employed in the pairwise potentials as shown in equation \ref{eq:Gaussian kernel}. $p_{i}$ and $p_{j}$ are the $x,y,z$ spatial information and $I_{i}$ and $I_{j}$ are the $R, G, B$ colour information. $\theta_{\alpha}, \theta_{\beta}$ and $\theta_{\gamma}$ are the parameters of Gaussian kernels.     

\begin{small}
\begin{equation}
k(f_i, f_j) = \omega^{(1)}exp(-\dfrac{|p_{i} - p_{j}|^2}{2\theta_{\alpha}^2}-\dfrac{|I_{i} - I_{j}|^2}{2\theta_{\beta}^2}) + \omega^{(2)}exp(-\dfrac{|p_{i} - p_{j}|^2}{2\theta_{\gamma}^2})
\label{eq:Gaussian kernel}
\end{equation}
\end{small}

Because fully-connected CRF considers the pairwise potentials over all pairs of pixels in the image, minimising the energy function exactly is intractable. Therefore, a mean-field approximation is employed for approximating maximum posterior marginal inference. The CRF distribution $P(X)$ is approximated by a simpler distribution $Q(X)$ by minimizing the KL-divergence $D(Q||P)$. This can be written as the product of independent marginal distributions, i.e. $ Q(X) = \prod_i Q_i(X_i)$. 

In CRF-RNN, one iteration of the mean-field algorithm can be formulated as a stack of common CNN layers, as shown in algorithm \ref{algorithm:mean-filed}. Multiple mean-field iterations can be implemented by repeating the above stack of layers. In other words, the repeated mean-field inference can be formulated as a Recurrent Neural Network(RNN). Then this CRF-RNN layer can be inserted as a part of the deep neural network after FCN-8s. During the training process, the error differentials of CRF-RNN can be passed to FCN-8s during backward propagation, so that the FCN-8s can generate better unary values for CRF-RNN optimization during the forward propagation. Meanwhile, the CRF parameters, such as the weights of the label compatibility function and Gaussian kernels, can be learned. 

\begin{algorithm}

$Q_{i} \leftarrow \dfrac{1}{Z_{i}} exp(U_{i}(l))$ for all $i$ 

\Comment{Initialization, {\color{red} $U$ from FCN-8s}}

\While{not converged} 
{ 
$ \tilde{Q}_{i}^{(m)}(l) \leftarrow \sum_{j\neq i}k^{(m)}(f_{i}, f_{j})Q_{j}(l)$ for all $m$ 

\Comment{Message Passing, {\color{red}Bilateral layer}}

$ \check{Q}_{i}(l) \leftarrow \sum_{m}w^{(m)}\tilde{Q}_{i}^{(m)}(l)  $

\Comment{Weighting Filter Outputs, {\color{red}Convolutional layer}}

$ \hat{Q}_{i}(l) \leftarrow \sum_{l^{'} \in L} \mu(l, l^{'})\check{Q}_{i}(l)  $

\Comment{Compatibility Transform, {\color{red}Convolutional layer}}

$ \breve{Q}_{i}(l) \leftarrow U_{i}(l) - \hat{Q}_{i}(l)  $

\Comment{Adding Unary Potentials, {\color{red}Concatenated layer}}

$ Q_{i} \leftarrow \dfrac{1}{Z_{i}}exp(\breve{Q}_{i}(l)) $

\Comment{Normalizing {\color{red}Softmax layer}}

}

\caption{Formulate one mean-field iteration as a stack of common CNN layers in \cite{Zheng2015}. The red annotations are the CNN layers related to corresponding steps in the mean-file iteration.}

\label{algorithm:mean-filed}

\end{algorithm}
\subsection{3D label refinement}
Following \cite{Hermans2014}\cite{McCormac2016}, Bayesian update is employed to fuse label hypotheses from the semantic point clouds in different views. Each voxel in a semantic point clouds stores the label information and the corresponding discrete probability. Using the camera projection and visual odometry, voxels from different viewpoints can be transformed to a common coordinate frame. This enables us to update the voxel's label probability distribution by means of a recursive Bayesian update, as shown in equation \ref{eq:Bayesian update}.
\begin{equation}
P(x = l_{i}|I_{1,...,k}) = \dfrac{1}{Z} P(x = l_{i}|I_{1,...,k-1})P(x = l_{i}|I_{k})
\label{eq:Bayesian update}
\end{equation}     

where $l_{i}$ is the label prediction, $I_{k}$ is the $k^{th}$ image and $Z$ is a constant for distribution normalization.
\section{Experiments}
In this section, we firstly introduce data preprocessing on a public material dataset MINC\cite{Bell2015}. Next the pipeline of network training is described. Finally, we present qualitative and quantitative evaluations of two different experiments: 2D material recognition on the MINC dataset, and 3D semantic reconstruction in a real-world application, respectively.

\subsection{Data preprocessing}
\begin{figure}[thpb]
	\raggedleft
	\centering
	\includegraphics[width=0.5\textwidth]{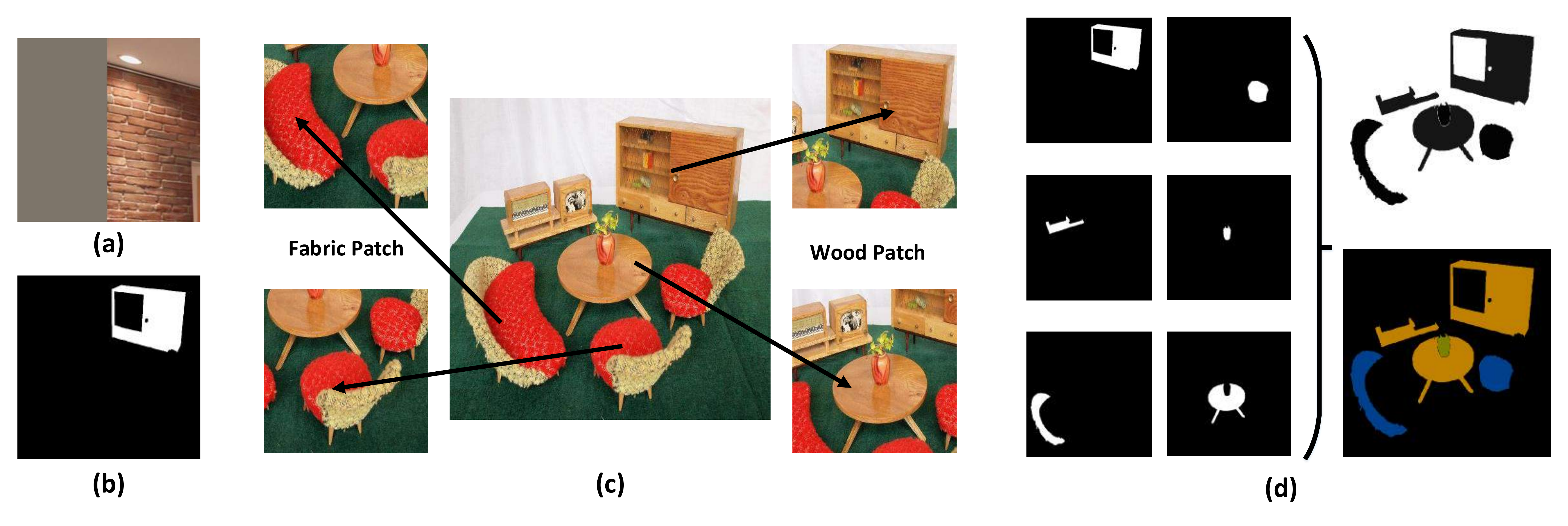}
	\caption{\textbf{Data preprocessing:} (a) A patch with non-values in MINC. (b) A partially pixel-wise labeled image in MINC. (c) Extracting new patches from original images in MINC. (d) Combining the partially pixel-wise labeled images to generate a fully pixel-wise labeled image.}
	\label{fig:Data_precessing}
\end{figure}

The large-scale public material dataset Materials in Context (MINC)\cite{Bell2015} is employed for training our neural network. MINC is diverse and well-sampled across 23 categories, including wood, glass, metal, brick, fabric and others. There are two kinds of human-annotated data in MINC: small RGB patches with a corresponding class labels(Fig.\ref{fig:Data_precessing}.(a)) and images which have been partially pixel-wise labeled at the object level(Fig.\ref{fig:Data_precessing}.(b)). 

Unfortunately, neither of these annotations can be used in our applications. In RGB patches, there are many non-values (e.g. grey parts in Fig.\ref{fig:Data_precessing}.(a)) because these patches extend beyond the limits of the image border. During the fine-tuning process, those non-values give a strong erroneous supervision to the VGG-16 network which lacks normalization layers. This prevents the VGG-16 network from converging. On the other hand, in the partially pixel-wise labeled images, all background pixels are masked, and only one foreground object is labeled, thus losing all context information for CRF-RNN training.

Some data preprocessing is required before network training. The original images from MINC are resized to 500$\times$500 images so that semantic segmentation, based on CNN, can be performed in real-time. Next, 256$\times$256 patches (which have one kind of material in the center) are extracted from the 500$\times$500 images, as shown in Fig.\ref{fig:Data_precessing}.(c). This ensures that there are no non-values in the extracted patches. 

Next, all partially pixel-wise labeled images, belonging to a single original image, are combined together to generate a single fully pixel-wise labeled image,  Fig.\ref{fig:Data_precessing}.(d). Because not all objects are labeled in the original image, it is not possible to generate 100\% pixel-wise labeled images. Therefore, any unlabeled pixels and repeated labeled pixels(object edges) are labeled as 255, which can then be ignored during the training process. Finally the pixel-wise labeled images are resized to 500$\times$500.  

82,1092 patches with the class labels for training, and 96,747 patches with the class labels for testing were generated. 1,498 pixel-wise labeled images for training and 300 pixel-wise labeled images for testing were generated.

\subsection{Network training}
We initialised our network weights using the publicly available weights of VGG-16\cite{Simonyan2015} model, pre-trained on ImageNet. However, the VGG-16 network is designed for the ImageNet challenge, which can classify 1000 different class labels. In contrast, for the MINC database, only 23 kinds of material need to be classified. Therefore we change the output number of inner-product layer fc8-minc to 23 instead of 1000. Next we fine-tune this network by using the newly extracted 256$\times$256 patches from MINC.

Because small patches have better spatial resolution, while large patch have more contextual information, the choice of patch size for fine-tuning is a trade-off between spatial resolution and contextual information. Four different patch sizes are tested in our fine-tuning experiments. As shown in Table \ref{table:patch_size_accuracy}, the accuracy of fine-tuning initially increases, but then decreases, as patch size grows. The highest (optimal) value is achieved when the patch size occupies around 30-50\% of the original image. 

\begin{table}[thpb]
\centering
\resizebox{\columnwidth}{!}{
\begin{tabular}{| c | c | c | c | c |}
\hline Patch size & 56$\times$56 & 156$\times$156 & 256$\times$256 & 356$\times$356 \\   
\hline Accuracy & 69.20\% & 81.06\% & 80.18\% & 73.40\% \\
\hline 
\end{tabular}}
\caption{\textbf{Accuracy of fine-tuning versus patch size.} Performance is best when the patch size occupies around 30-50\% of the original image.}
\label{table:patch_size_accuracy}
\end{table}       

After fine-tuning, the fully connected VGG-16 network is transplanted to a fully convolutional VGG-16 network. The last three inner product layers (fc6, fc7 and fc8-minc) are transformed to the convolutional layers (fc6-conv, fc7-conv and fc8-conv). Convolutional layers inherit the weights of the inner product layers. 

Using the fine-tuning model, the FCN-32s, FCN-16s and FCN-8s are trained step by step using 1,498 fully pixel-wise labeled images. Next, the CRF-RNN layer is inserted to form the part of the network following the FCN-8s. After inheriting the learned weights, this end-to-end FCN-8s with CRF-RNN network is trained again using 1,498 fully pixel-wise labeled images. The parameters of each trained network are shown in Table \ref{table:parameter of network}. The number of mean-field iterations \textit{T} in the CRF-RNN is set to 5 during the training process. This helps avoid vanishing gradient problems and reduces the training time. During the test process, the number of mean-field iterations can be kept at 5 or be increased to 10 depending on the run-time required.   

\newcommand{\tabincell}[2]{\begin{tabular}{@{}#1@{}}#2\end{tabular}}
\begin{table}[thpb]
\centering
\resizebox{\columnwidth}{!}{
\begin{tabular}{| c | c | c | c | c | c |}
\hline             & Learning rate & Momentum & Batch size &Weight decay & Training data \\   
\hline Fine-tuning & \tabincell{c}{1e-4 reduction \\ with 0.1} & 0.95 & 50 & 0.0005 & \tabincell{c}{256*256 \\ RGB patch} \\
\hline FCN-32s     & 1e-10 & 0.99 & 1 & 0.0005 & \tabincell{c}{500*500 \\ RGB image} \\
\hline FCN-16s     & 1e-12 & 0.99 & 1 & 0.0005 & \tabincell{c}{500*500 \\ RGB image} \\
\hline FCN-8s      & 1e-14 & 0.99 & 1 & 0.0005 & \tabincell{c}{500*500 \\ RGB image} \\
\hline \tabincell{c}{FCN-8s with \\ CRF-RNN}     & 1e-12 & 0.99 & 1 & 0.0005 & \tabincell{c}{500*500 \\ RGB image} \\
\hline 
\end{tabular}}
\caption{The parameters of trained network.}
\label{table:parameter of network}
\end{table}           
\subsection{2D material recognition}
We evaluated our trained network using 300 fully pixel-wise labeled images from MINC for 2D pixel-wise material recognition. 

\subsubsection{The qualitative analysis}
\begin{figure}[thpb]
\centering
\subfigure[\label{fig:test1}]{\includegraphics[width= 0.11\textwidth]{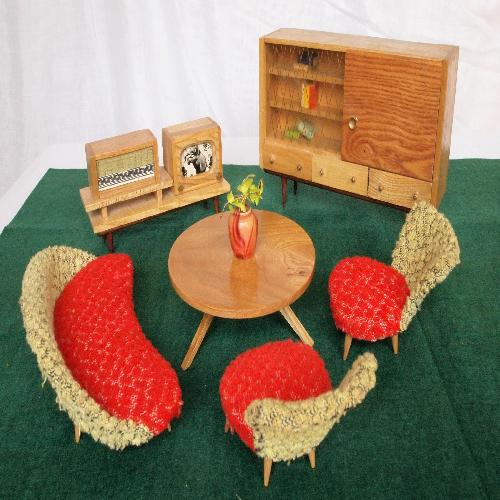}}
\subfigure[\label{fig:test2}]{\includegraphics[width= 0.11\textwidth]{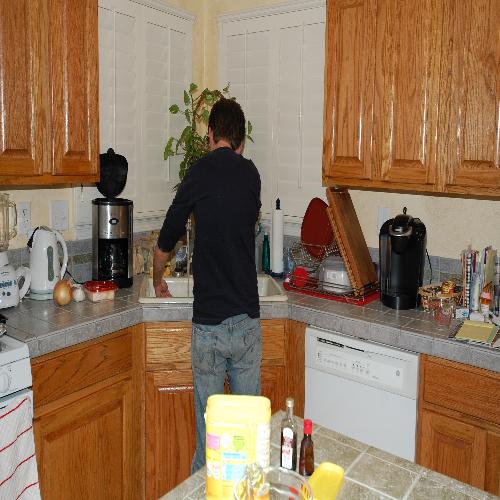}}
\subfigure[\label{fig:test3}]{\includegraphics[width= 0.11\textwidth]{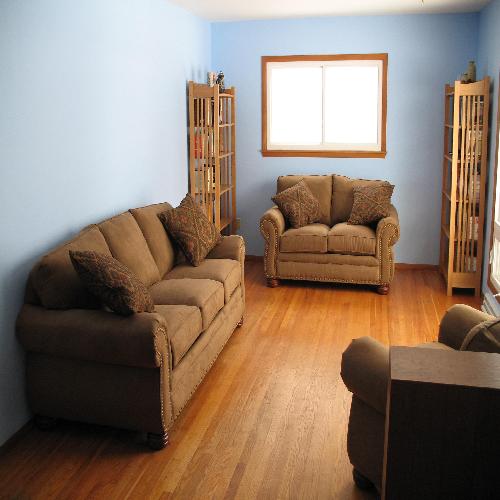}}
\subfigure[\label{fig:test4}]{\includegraphics[width= 0.11\textwidth]{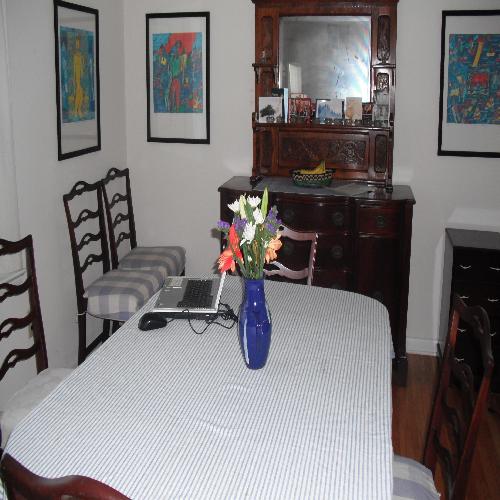}}

\subfigure[\label{fig:groundtruth_1}]{\includegraphics[width= 0.11\textwidth]{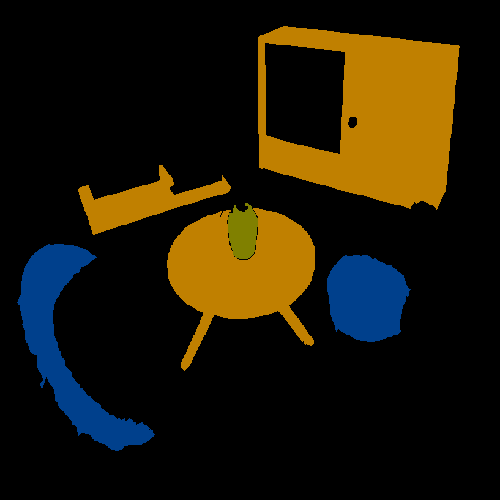}}
\subfigure[\label{fig:groundtruth_2}]{\includegraphics[width= 0.11\textwidth]{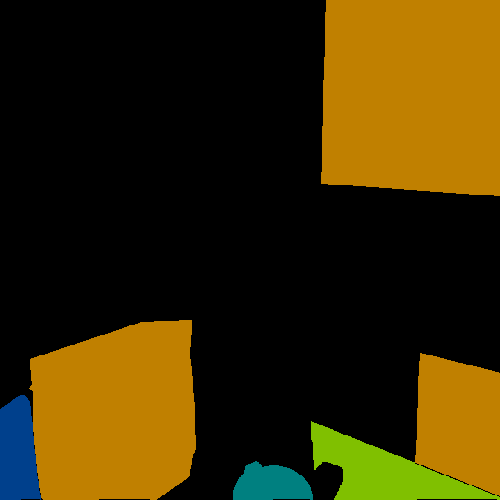}}
\subfigure[\label{fig:groundtruth_3}]{\includegraphics[width= 0.11\textwidth]{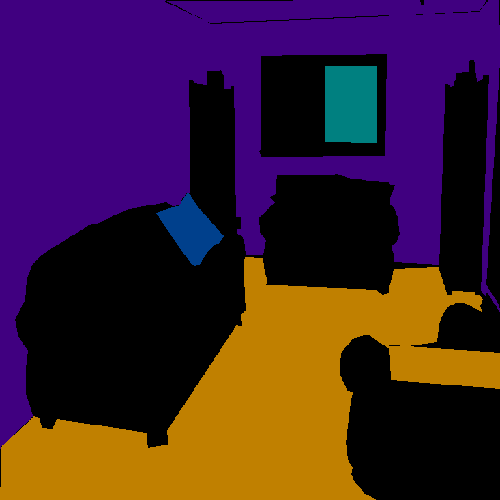}}
\subfigure[\label{fig:groundtruth_4}]{\includegraphics[width= 0.11\textwidth]{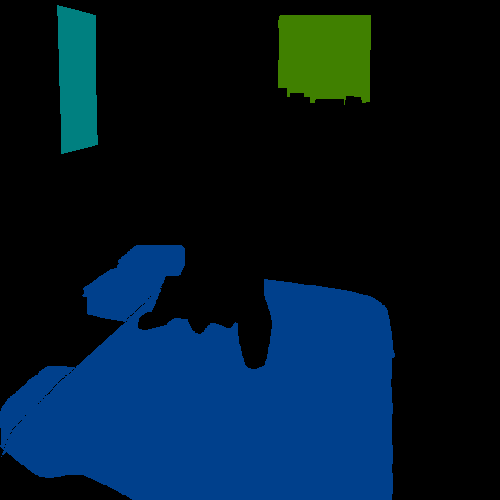}}

\subfigure[\label{fig:test1_FCN8}]{\includegraphics[width= 0.11\textwidth]{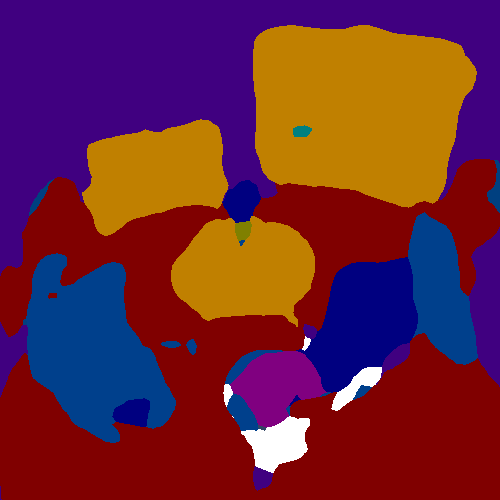}}
\subfigure[\label{fig:test2_FCN8}]{\includegraphics[width= 0.11\textwidth]{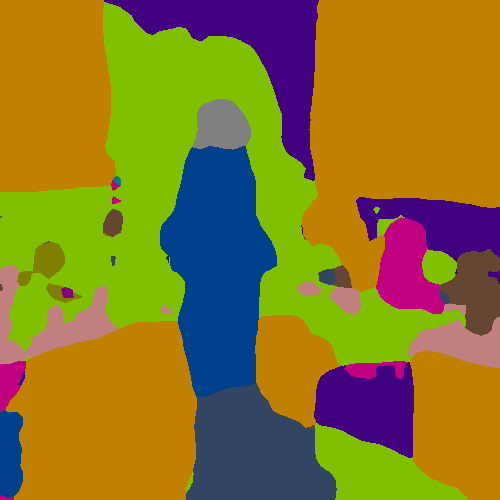}}
\subfigure[\label{fig:test3_FCN8}]{\includegraphics[width= 0.11\textwidth]{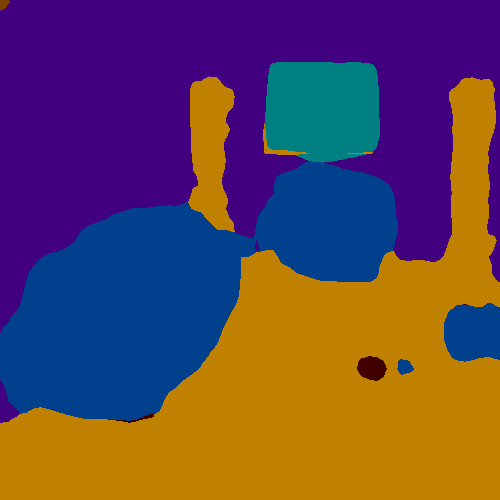}}
\subfigure[\label{fig:test4_FCN8}]{\includegraphics[width= 0.11\textwidth]{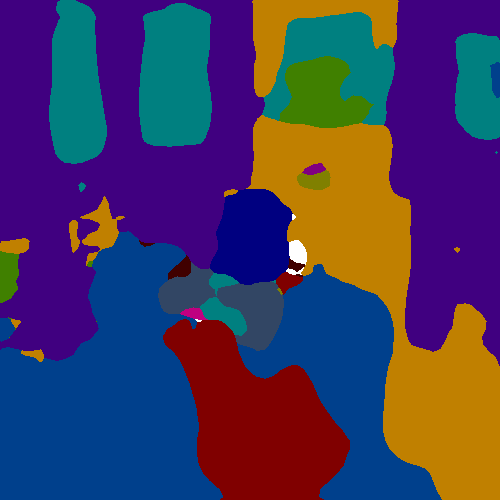}}

\subfigure[\label{fig:test1_CRFRNN}]{\includegraphics[width= 0.11\textwidth]{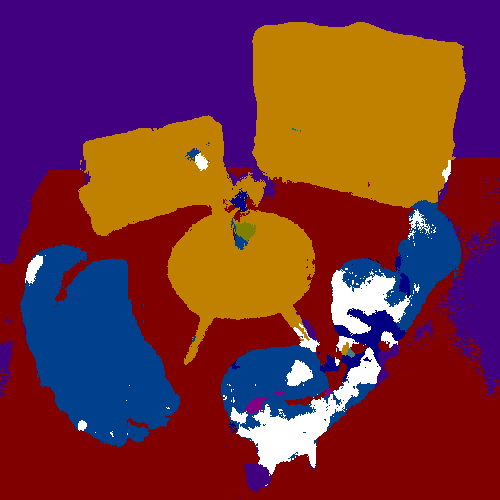}}
\subfigure[\label{fig:test2_CRFRNN}]{\includegraphics[width= 0.11\textwidth]{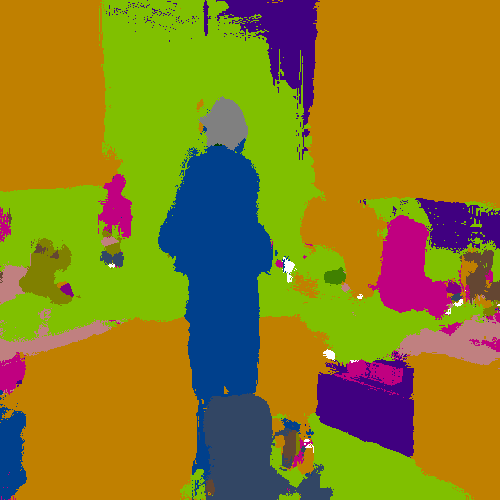}}
\subfigure[\label{fig:test3_CRFRNN}]{\includegraphics[width= 0.11\textwidth]{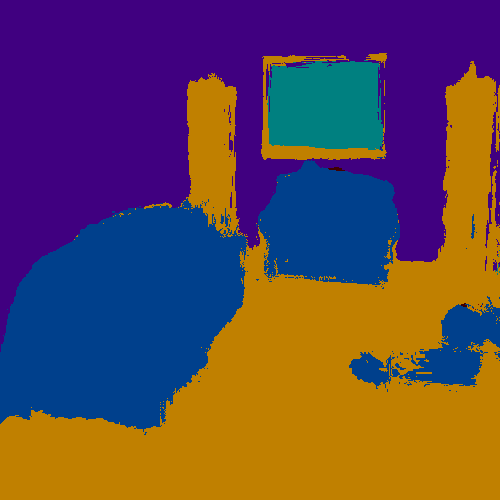}}
\subfigure[\label{fig:test4_CRFRNN}]{\includegraphics[width= 0.11\textwidth]{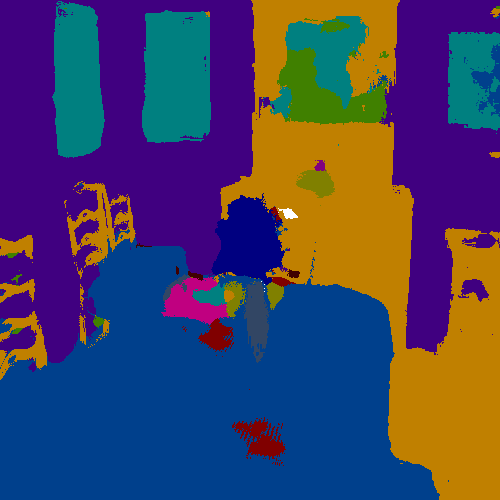}}

\caption{\textbf{Qualitative results of 2D material recognition in MINC:} (a)(b)(c)(d) are original RGB images, (e)(f)(g)(h) are ground truth images, (i)(j)(k)(l) are semantic segmentation results of FCN-8s, (m)(n)(o)(p) are semantic segmentation results of FCN-8s with CRF-RNN. Clearly the semantic results of FCN-8s with CRF-RNN generate much clearer shapes than FCN-8s alone, e.g. table leg in (m), person in (n), sofa in (o), and chair back and vase in (p). In (l), a large part of fabric is erroneously labeled as carpet, while the size of this error greatly decreases in (P) due to the neighbourhood consistency constraints of the fully connected CRF optimization.}
\label{fig:qualitative_CRFRNN}
\end{figure}

The semantic segmentation results of FCN-8s have non-sharp boundaries because of lacking neighbourhood consistency constraints. After inserting CRF-RNN into the network after FCN-8s, semantic label assignments are significantly improved. As shown in Fig.\ref{fig:qualitative_CRFRNN}, the first and second rows are the original and ground-truth images in MINC. The third and fourth rows are the 2D semantic segmentation results of FCN-8s, and FCN-8s with CRF-RNN, respectively. Clearly the semantic results of FCN-8s with CRF-RNN generate much clear shapes than FCN-8s alone, e.g. table leg in (m), person in (n), sofa in (o), and the chair back and vase in (p). In (l), a large section of ``fabric'' is erroneously recognised as ``carpet''. In contrast, this erroneous section is much smaller in (P) because of the neighbourhood consistency constraints of the fully connected CRF optimization.

\subsubsection{Quantitative analysis}
\begin{figure}[thpb]
\centering
\subfigure[\label{fig:FCN8_confusion_matrix_1}]{\includegraphics[width=0.35\textwidth]{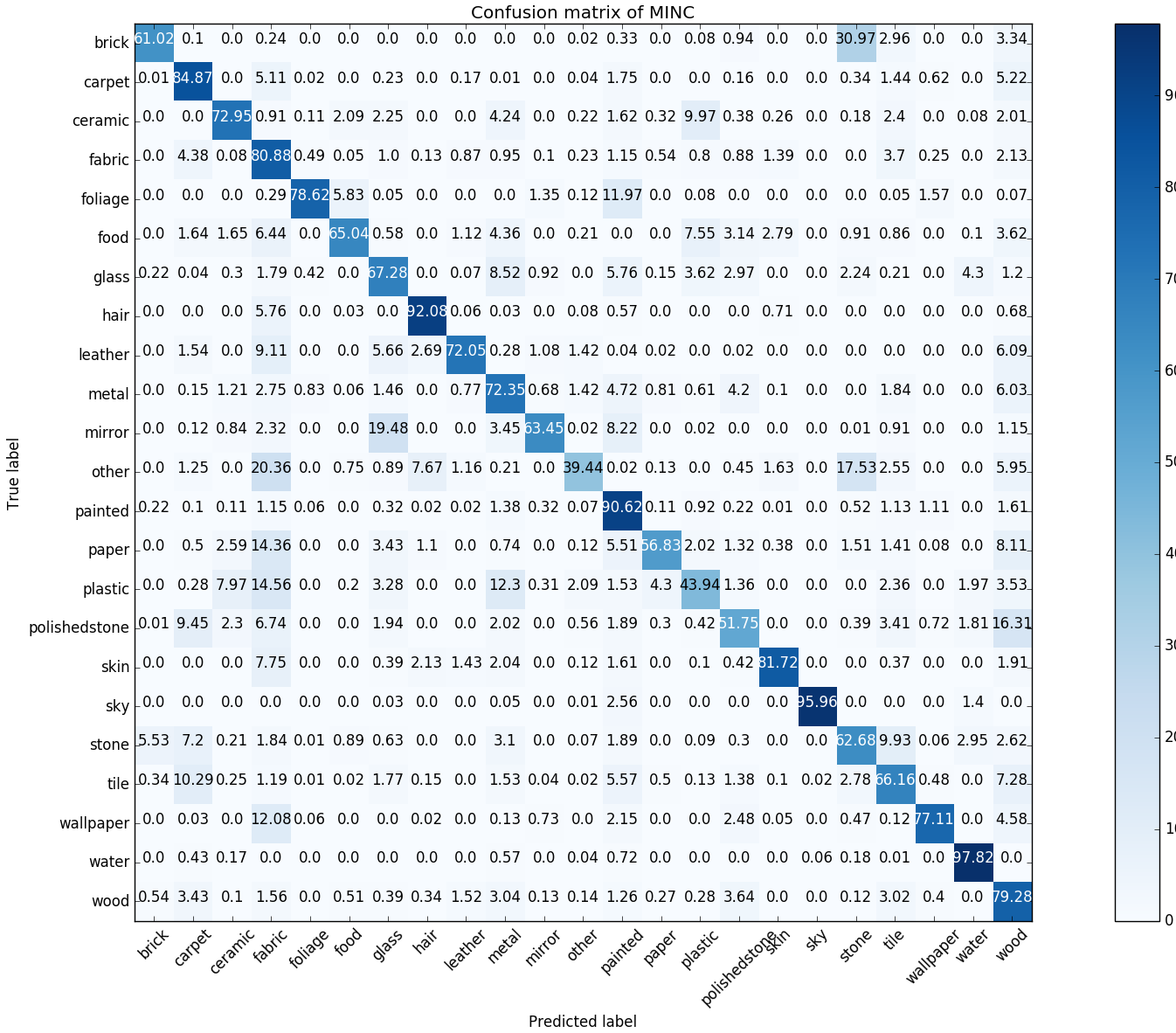}}
\subfigure[\label{fig:CRFRNN_confusion_matrix1}]{\includegraphics[width=0.35\textwidth]{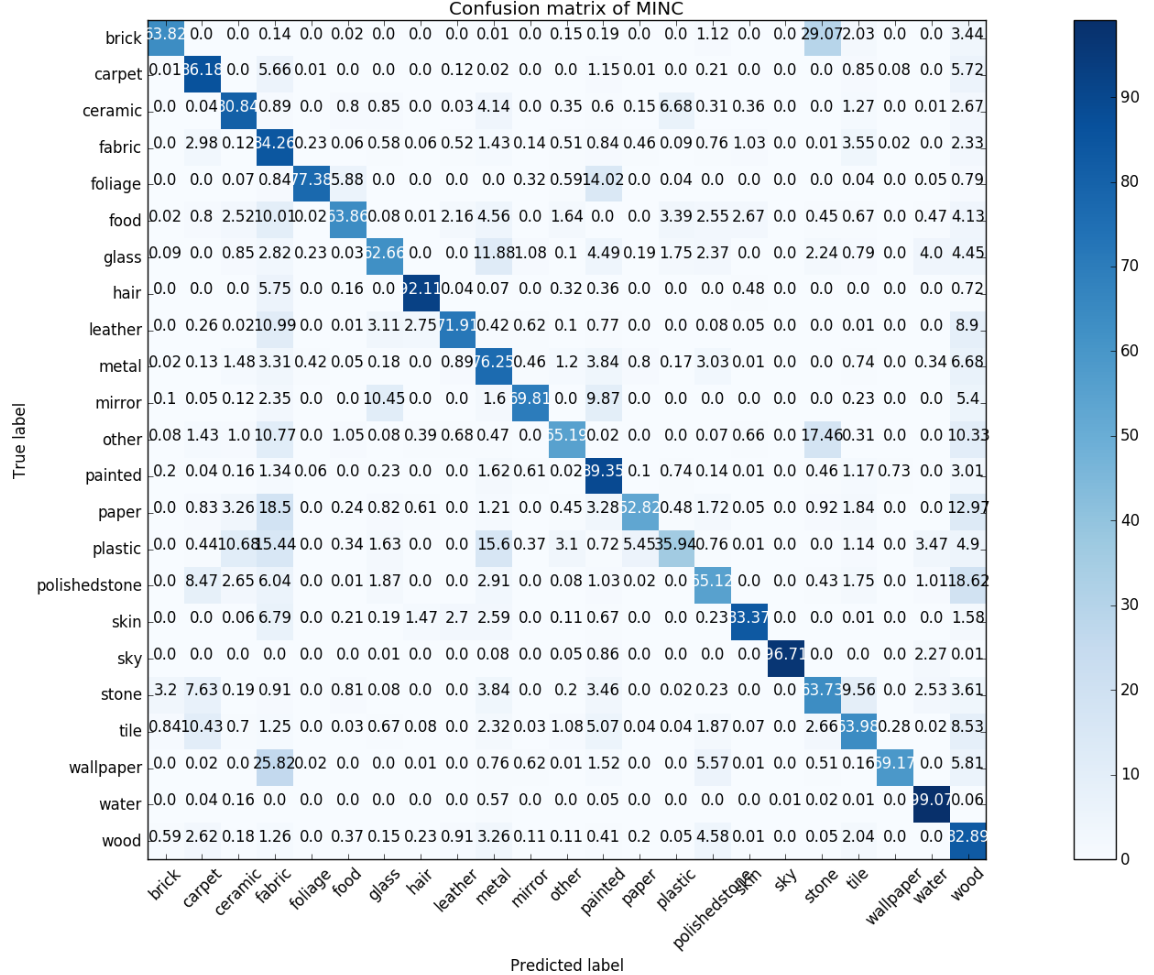}}
\caption{\textbf{Quantitative results of 2D material recognition in MINC.} Confusion matrices of (a)FCN-8s and (b)FCN-8s with CRF-RNN. The colour in the diagonal line is much darker than that in the other positions, suggesting good performance. After combining CRF-RNN with FCN-8s in a united framework, the recognition rate of each class increases by around 4-6\%. }
\label{fig:confusion_matrix}
\end{figure}

The standard parameters for scene understanding evaluation: \textit{pixel accuracy, mean accuracy, mean IU} and \textit{frequency weighed IU} are used for quantitative analysis, as shown in Table.\ref{table:2D Quantitative results}. End-to-end FCN-8s with CRF-RNN improve 3.53\%, 5.16\%, 4.62\% and 3.92\% for \textit{pixel accuracy, mean accuracy, mean IU} and \textit{frequency weighed IU} respectively, as compared to FCN-8s without CRF-RNN. The confusion matrices of material recognition are shown in Fig.\ref{fig:confusion_matrix}. The colour in the diagonal line is much darker than that in the other positions, suggesting good performance. After combining CRF-RNN with FCN-8s in a united framework, the recognition rate of each class increases by around 4-6\%. We attribute this relatively small improvement to the small number (300) of only partially labeled images for this test.


\begin{table}[thpb]
\centering
\resizebox{\columnwidth}{!}{
\begin{tabular}{| c | c | c | c | c |}
\hline          & Pixel acc. & Mean acc. & Mean IU & f.w. IU \\   
\hline FCN-8s   & 78.41\% & 71.91\% & 56.51\% & 66.07\% \\
\hline \tabincell{c}{FCN-8s with \\ CRF-RNN} & 81.94\% & 77.07\% & 61.13\% & 69.99\% \\
\hline 
\end{tabular}}
\caption{\textbf{Quantitative results of 2D material recognition in MINC.} End-to-end FCN-8s with CRF-RNN improve 3.53\%, 5.16\%, 4.62\% and 3.92\% for \textit{pixel accuracy, mean accuracy, mean IU} and \textit{frequency weighed IU} respectively, compared with FCN-8s alone. }
\label{table:2D Quantitative results}
\end{table} 

\subsubsection{Run-time performance}
Our experiments were performed using an i7-6800k(3.4Hz) 8-cores CPU and NVIDIA TITAN X GPU (12G). For a 500$\times$500 image, the 2D semantic segmentation based on the GPU version of FCN-8s costs 0.13s-0.15s, and that of FCN-8s with CRF-RNN costs 0.4s-0.6s (10 iterations) or 0.2s-0.3s (5 iterations). The run-time greatly decreases if smaller RGB images, e.g. 224$\times$224, are used, enabling real-time, or near-to-real-time, pixel-wise material segmentation.  
\subsection{3D semantic reconstruction}
We next evaluate our proposed method in a real-world application. Simultaneous 3D reconstruction and material recognition was performed in a real office which contains many different materials, such as brick, wood, metal, paper, carpet, painted surfaces, and others. 

\subsubsection{Quantitative analysis}
\begin{figure}[thpb]
\centering
\subfigure[\label{fig:rgb1}]{\includegraphics[width= 0.2\textwidth]{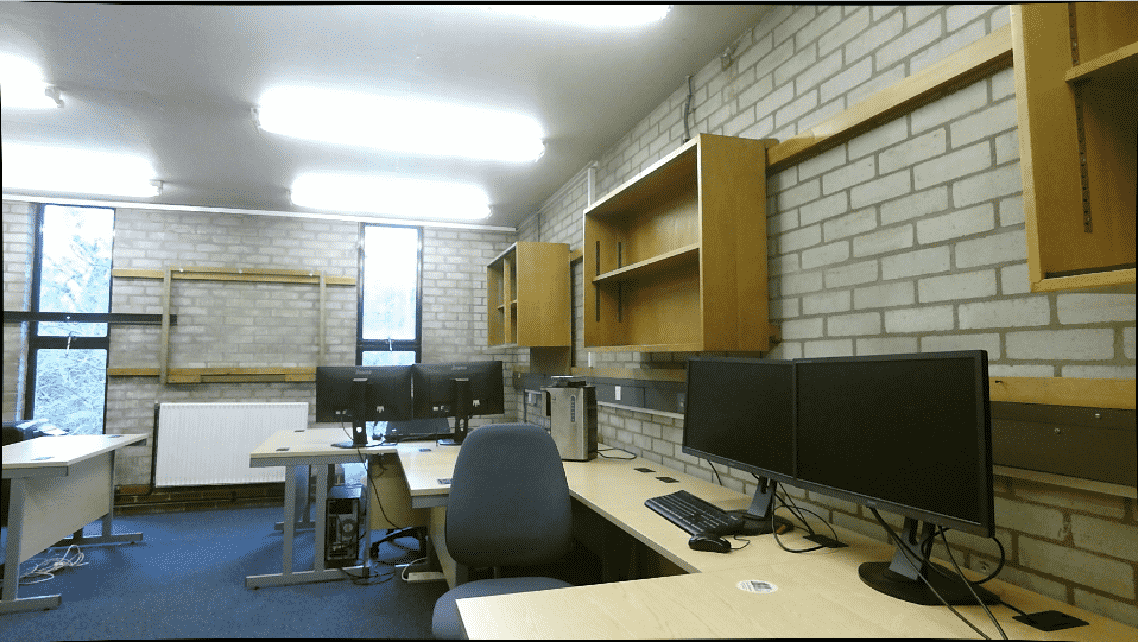}}
\subfigure[\label{fig:rgb2}]{\includegraphics[width= 0.2\textwidth]{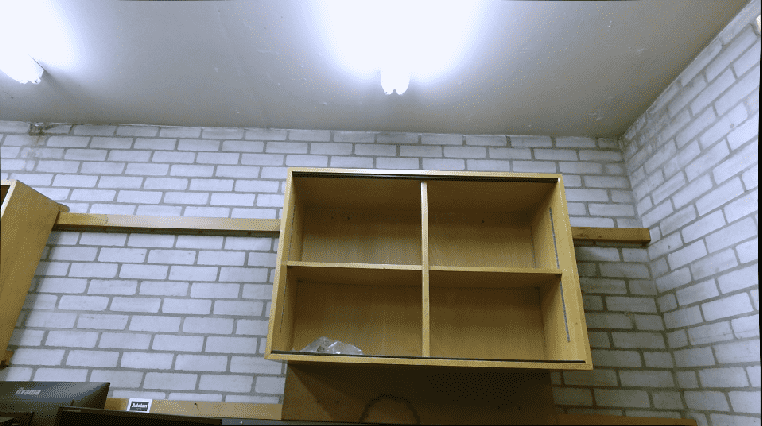}}

\subfigure[\label{fig:semantic1}]{\includegraphics[width= 0.2\textwidth]{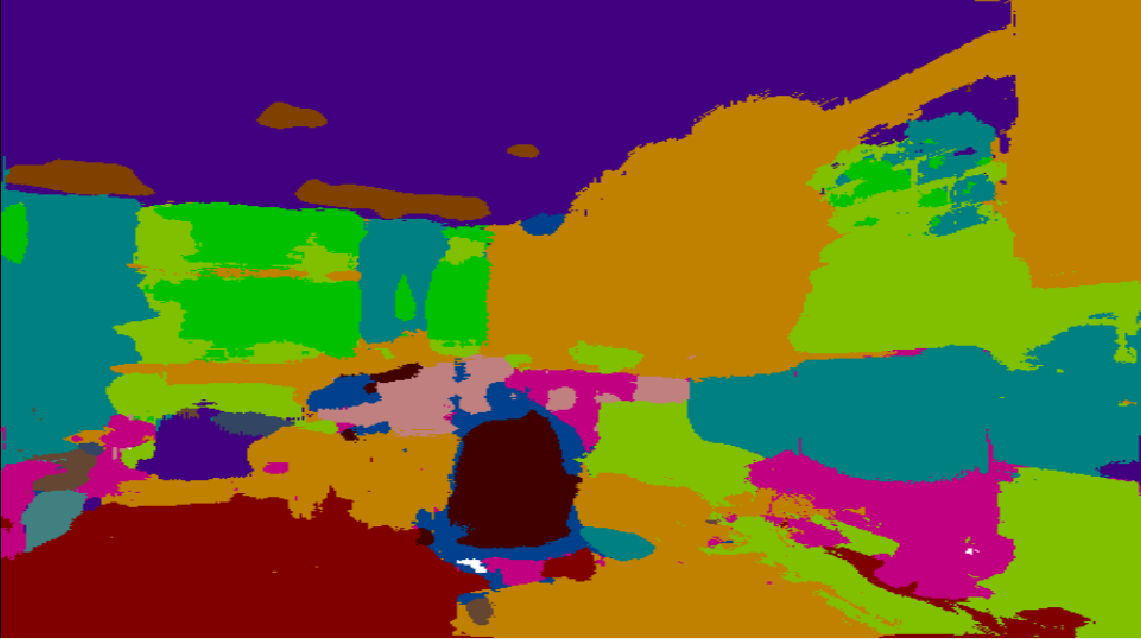}}
\subfigure[\label{fig:semantic2}]{\includegraphics[width= 0.2\textwidth]{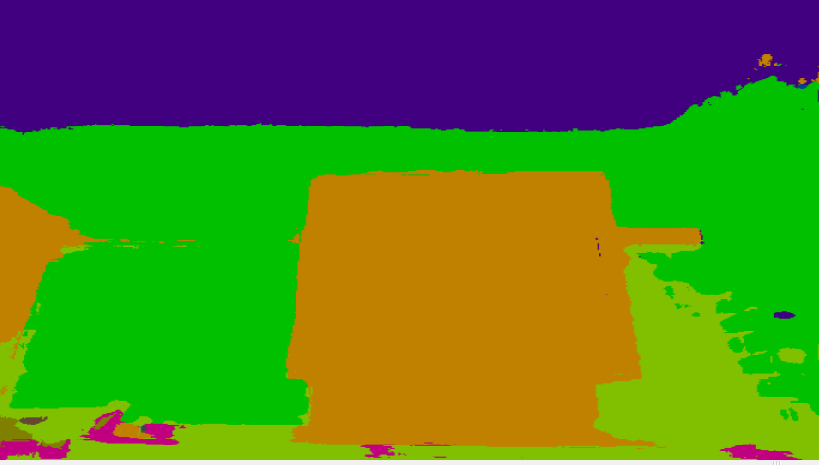}}

\subfigure[\label{fig:pointcloud1}]{\includegraphics[width= 0.2\textwidth]{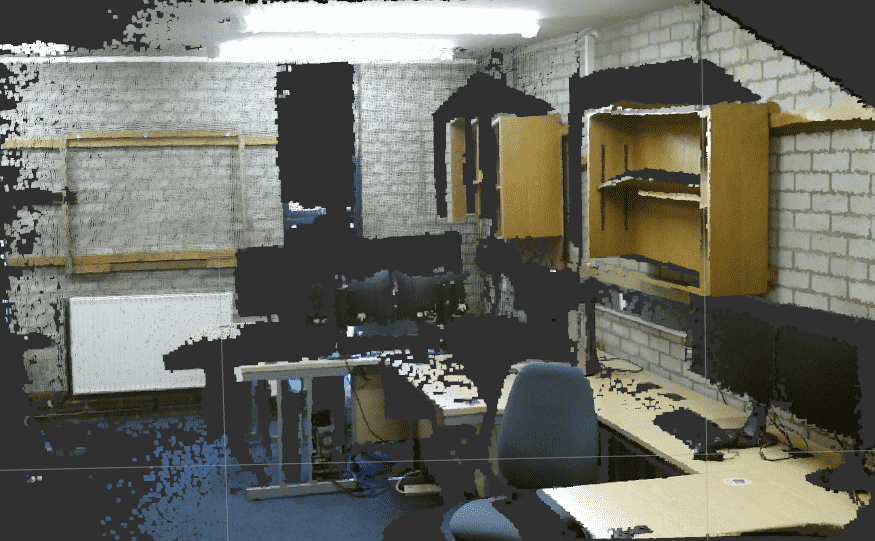}}
\subfigure[\label{fig:pointcloud2}]{\includegraphics[width= 0.2\textwidth]{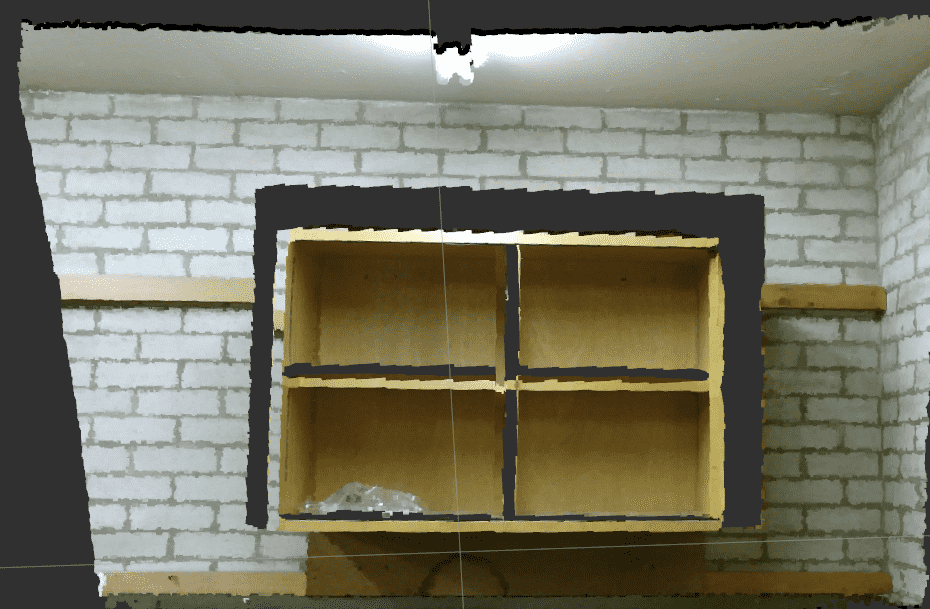}}

\subfigure[\label{fig:semantic_pointcloud1}]{\includegraphics[width= 0.2\textwidth]{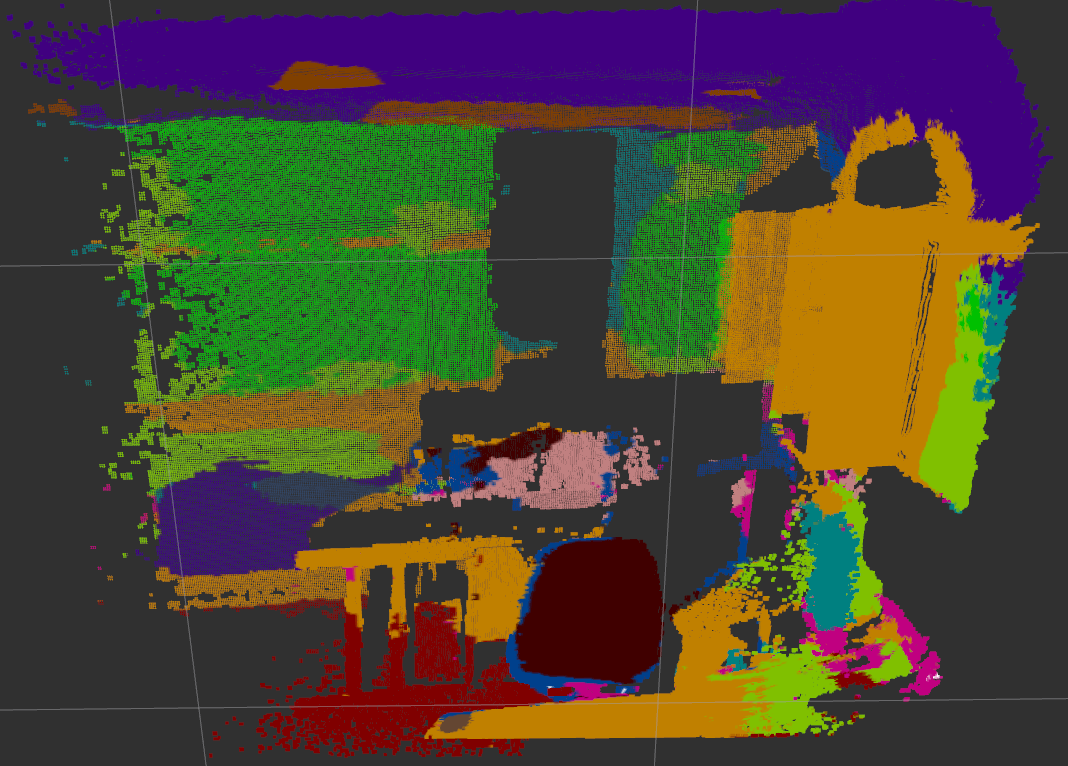}}
\subfigure[\label{fig:semantic_pointcloud2}]{\includegraphics[width= 0.2\textwidth]{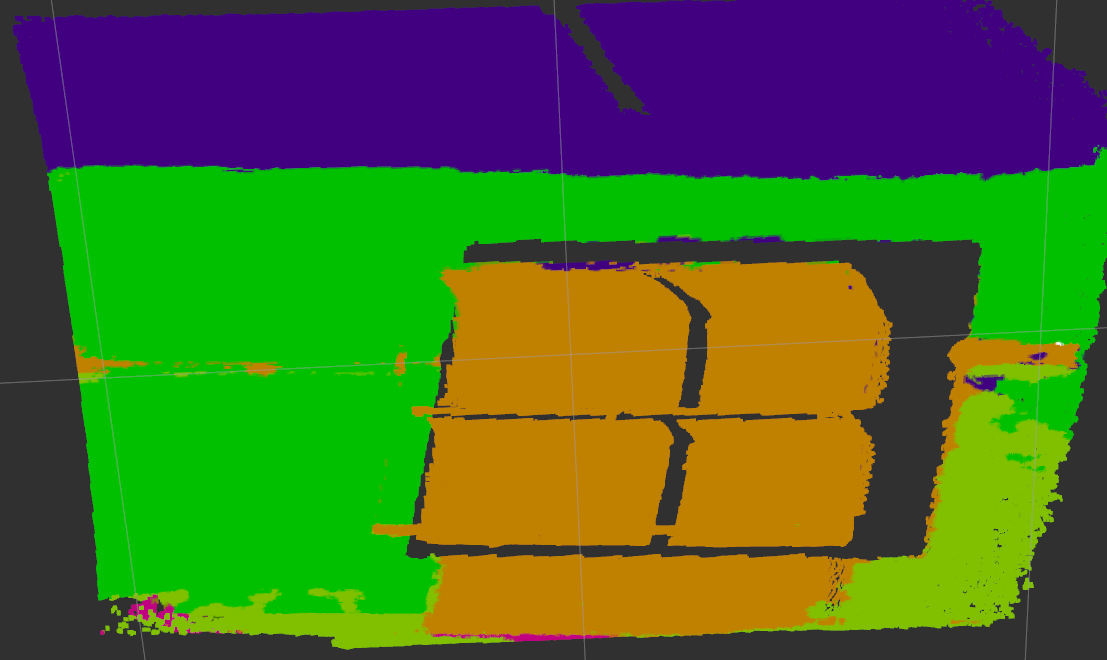}}

\caption{\textbf{Qualitative results of 3D semantic reconstruction in a multi-material office}: RGB images from Kinect2 (a)(b), 2D semantic segmentation images (c)(d), 3D point clouds (e)(f), 3D semantic point clouds (g)(h)}
\label{fig:semantic_pointcloud_experiment1}
\end{figure}

\begin{figure}[thpb]
\centering
\subfigure[\label{fig:pointcloud_wall}]{\includegraphics[width= 0.4\textwidth]{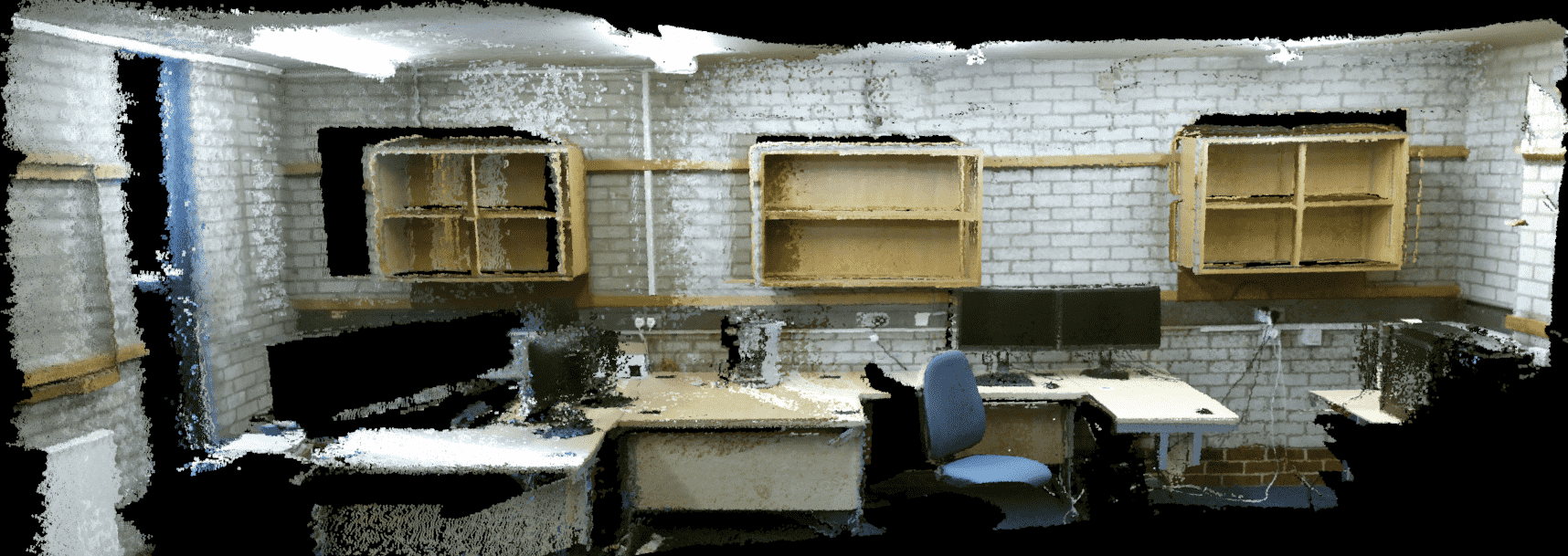}}
\subfigure[\label{fig:semantic_pointcloud_wall}]{\includegraphics[width= 0.4\textwidth]{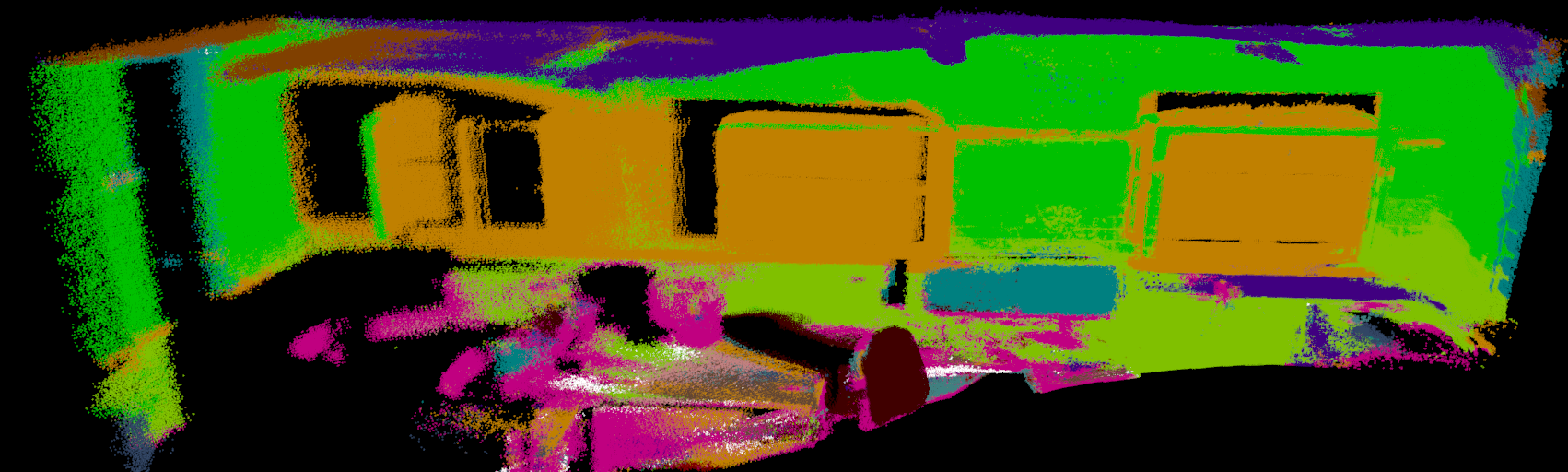}}
\caption{\textbf{Qualitative results of 3D semantic reconstruction in a multi-material office:} (a) Local 3D map. (b) Local 3D semantic map.}
\label{fig:semantic_pointcloud_experiment2}
\end{figure}

\begin{figure}[thpb]
\centering
\subfigure[\label{fig:3D_reconstruction_lab}]{\includegraphics[width= 0.3\textwidth]{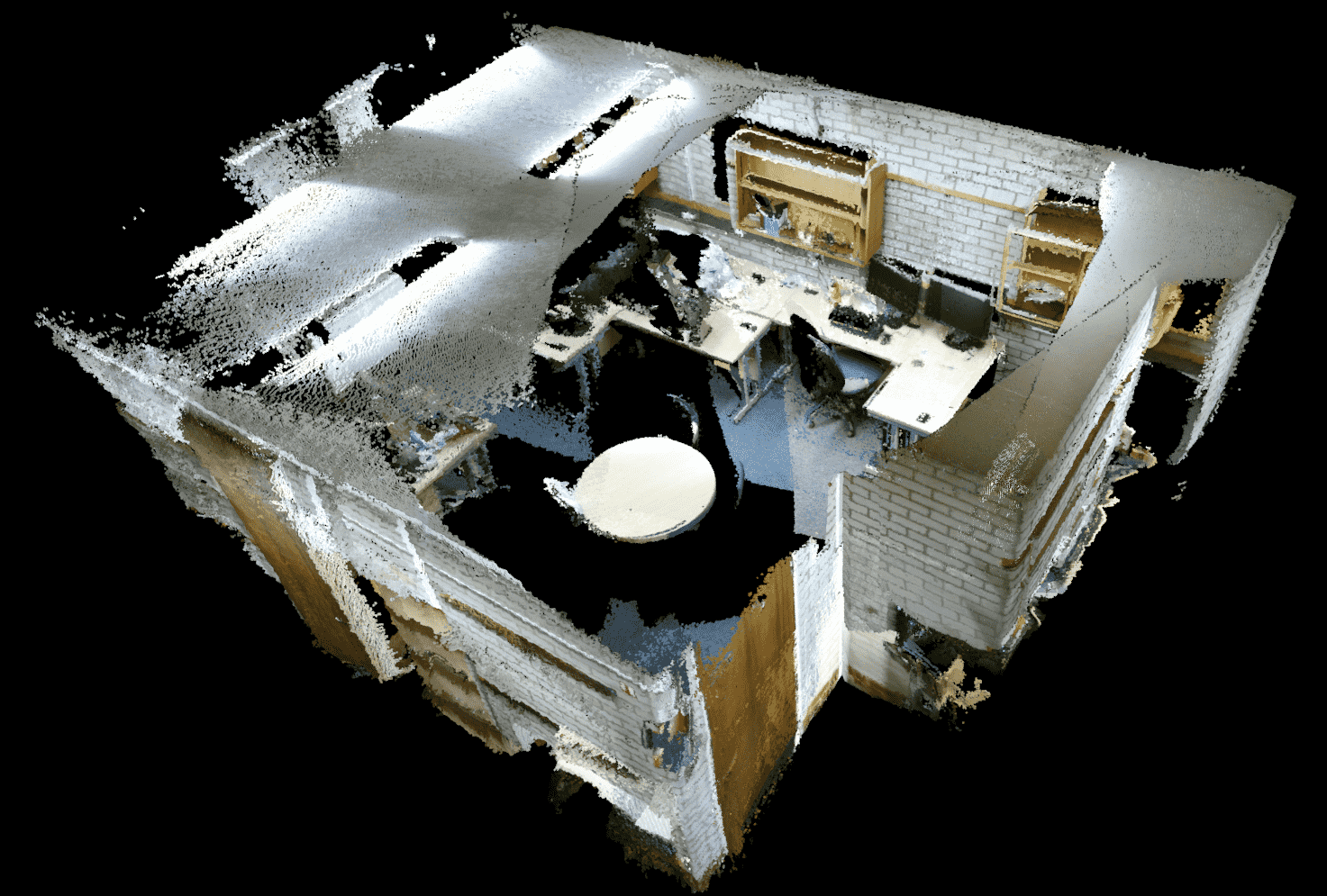}}
\subfigure[\label{fig:semantic_room}]{\includegraphics[width= 0.3\textwidth]{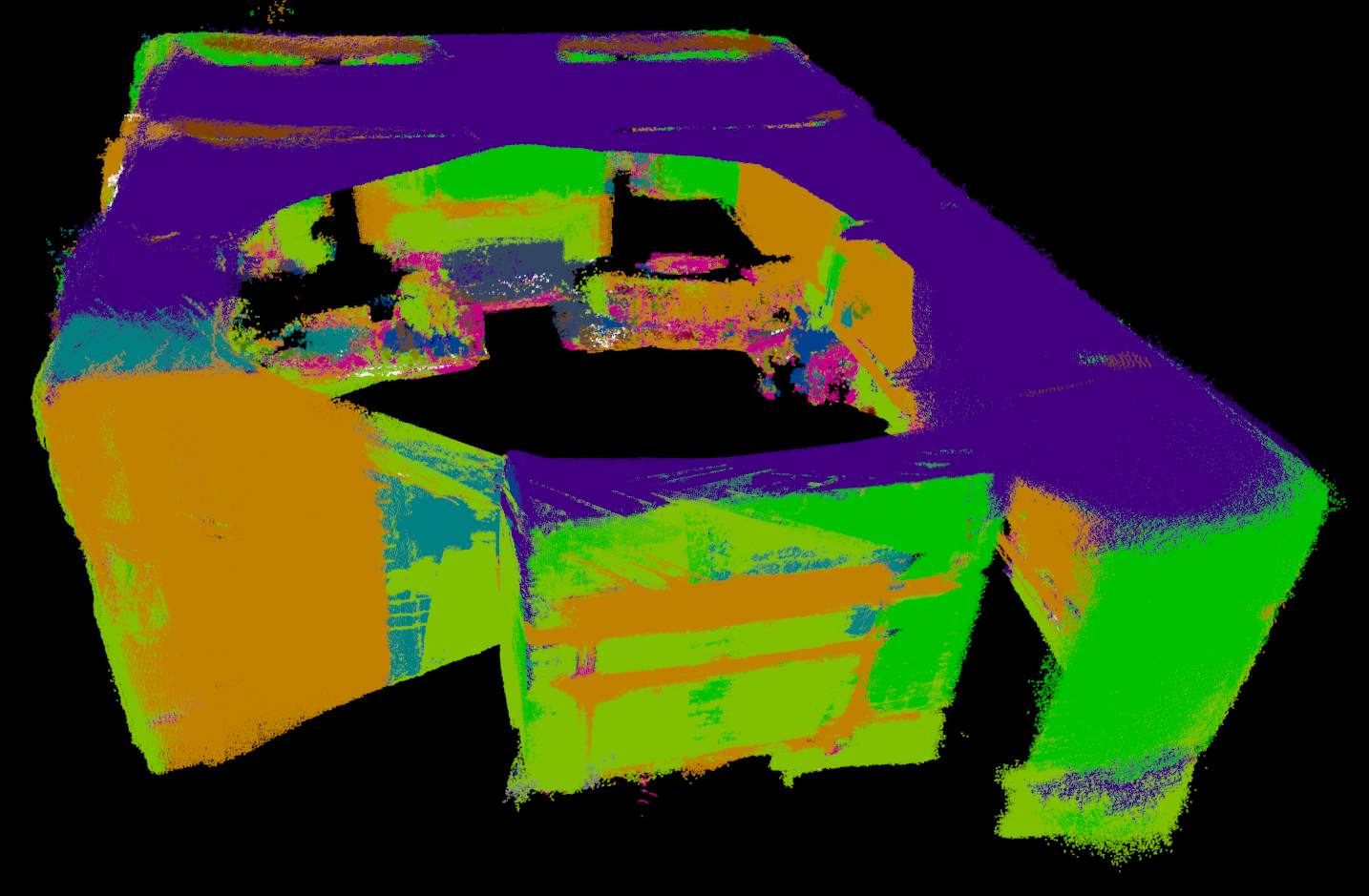}}
\caption{\textbf{Qualitative results of 3D semantic reconstruction in a multi-material office:} (a) Global 3D map. (b) Global 3D semantic map.}
\label{fig:semantic_pointcloud_experiment3}
\end{figure}

The qualitative results of each step in our system in a multi-material office are shown in Fig.\ref{fig:semantic_pointcloud_experiment1}. The local/global 3D map and local/global 3D semantic map are shown in Fig. \ref{fig:semantic_pointcloud_experiment2} and \ref{fig:semantic_pointcloud_experiment3} respectively. It can be seen that most of materials are correctly classified and segmented. However, some small objects cannot be recognised because they do not provide enough pixels in the RGB image. The pixel in the border between two different materials is easily assigned a wrong prediction label. In addition, some errors also result from illumination variances.   

\subsubsection{Quantitative analysis}
\textit{Pixel accuracy, mean accuracy, mean IU and frequency weighed IU} are used for quantitative evaluation. First, 40 key frames of 3D reconstruction in the multi-material office were obtained according to visual odometry from RGBD SLAM. Next, we densely annotated all materials in all key frames using JS Segment Annotator\footnote{http://kyamagu.github.io/js-segment-annotator/}. Finally, pixel-wise true or false numbers were counted between the corresponding pixels from ground-truth and predicted images. 

Table \ref{table:3D Quantitative results} shows quantitative results. \textit{Pixel accuracy} (80.10\%) is satisfactory, while \textit{mean accuracy} (58.75\%) appears much lower than that reported in MINC evaluation (76.87\%). However, these numbers are misleading, because we only use 40 test samples and there is large variance in material detection rates. Pixel-wise recognition accuracy of some materials e.g. mirror(0\%) and paper(6.78\%) is very low. However, the mirror only appears in one instance. So just one failure to recognise mirror generates a score of 0\%, which greatly decreases the overall mean accuracy score. In future, we will obtain more samples, in different scenes, to perform more informative quantitative analysis.

\begin{table}[thpb]
\centering
\resizebox{\columnwidth}{!}{
\begin{tabular}{| c | c | c | c | c |}
\hline          & Pixel acc. & Mean acc. & Mean IU & f.w. IU \\   
\hline \tabincell{c}{3D semantic \\ reconstruction}  & 80.10\% & 58.75\% & 39.45\% & 68.76\% \\
\hline 
\end{tabular}}
\caption{\textbf{Quantitative results of 3D semantic reconstruction in a multi-material office.} The \textit{pixel accuracy}(80.10\%) is good, while the \textit{mean accuracy}(58.75\%) appears much lower than that(76.87\%) of MINC evaluation. However, this is an artifact of the small number of samples (40). Failure to detect just one instance of mirror, causes an accuracy of 0\% for that category, which misleadingly skews the overall mean accuracy score to appear low.}
\label{table:3D Quantitative results}
\end{table} 

\subsubsection{Implementation and run-time performance}  
We implemented our system on an i7-6800k(3.4Hz) 8-cores CPU and NVIDIA TITAN X GPU (12G). IAI Kinect2 package\footnote{https://github.com/code-iai/iai\_kinect2/} is employed to interface with ROS and calibrate the Kinect's RGB and depth cameras. The FCN-8s with CRF-RNN is implemented using Caffe toolbox\footnote{http://caffe.berkeleyvision.org/}. The overall system is implemented using C++ and GPU programming within a ROS framework. 

Run-time performance of our system is around 2Hz (10 iterations) or 4Hz (5 iterations) using the QHD RGB and depth images from Kinect2. The 540$\times$960 RGB image is reduced to 500$\times$500 RGB image for material recognition, and then increased to 540$\times$960 RGB image for semantic reconstruction.
The run-time performance can be boosted to around 10Hz if the QHD RGB image is decreased to 224$\times$224 RGB image, using 5 CRF iterations for material recognition. In contrast, the run-time performance of SemanticFusion\cite{McCormac2016} claims up to 25.3Hz using 224$\times$224 RGB image. However, this does not include the significant time needed for CRF post-processing. On average, SemanticFusion takes 20.3s to perform 10 CRF iterations. In contrast, our system is a fully end-to-end system and our run-time includes the CRF optimization which is embedded within our network. For real-time 3D reconstruction, most of the frames are abandoned and only a few key frames are used. So 5Hz-10Hz run-time performance is enough to ensure a real-time semantic reconstruction assuming a 30fps RGB-D camera.

A video demo can be found \url{https://www.youtube.com/watch?v=bVbrb_aE6uw}.
\section{Conclusions}
In this paper, we report the first system for simultaneous 3D reconstruction and material recognition. It is a real-time, fully end-to-end system, which does not require hand-crafted features or post-processing CRF optimization. Its run-time performance can be boosted to around 10Hz, enabling real-time 3D semantic reconstruction with a 30fps camera. We presented both quantitative and qualitative experimental results, which support the effectiveness of our method.

\section*{Acknowledgment}
This work was supported by H2020 RoMaNS and 645582, EPSRC grant EP/M026477/1. Zhao was supported by DISTINCTIVE scholarship. Sun was support by RoMaNS. Stolkin was supported by a Royal Society Industry Fellowship.
\bibliographystyle{IEEEtran}
\bibliography{./library.bib}
\end{document}